%% file: root.tex
\documentclass[letterpaper, 10 pt, conference]{ieeeconf}  

\IEEEoverridecommandlockouts                              

\title{\LARGE \bf
End-to-End Crop Row Navigation via LiDAR-Based Deep Reinforcement Learning
}

\author{Ana Luiza Mineiro$^{*}$, Francisco Affonso$^{*}$ and Marcelo Becker
\thanks{*These authors contributed equally.}%
\thanks{The publication was written prior to Francisco Affonso joining University of Illinois Urbana Champaign (UIUC).}
\thanks{All authors are with the Department of Mechanical Engineering, University of São Paulo (USP), BR {\tt\small \{analuizamineiro, faffonso\}@usp.br, becker@sc.usp.br}}%
}

\usepackage{graphicx}
\usepackage{pgf}
\usepackage{pgfplots}
\usepackage{svg}
\usepackage{amsfonts}
\usepackage{amsmath}
\usepackage{bbm}
\usepackage{pgffor}
\usepackage{booktabs}
\usepackage{cite}
\usepackage{dsfont}

\begin{document}

\maketitle
\thispagestyle{empty}
\pagestyle{empty}

\begin{abstract}

Reliable navigation in under-canopy agricultural environments remains a challenge due to GNSS unreliability, cluttered rows, and variable lighting. To address these limitations, we present an end-to-end learning-based navigation system that maps raw 3D LiDAR data directly to control commands using a deep reinforcement learning policy trained entirely in simulation. Our method includes a voxel-based downsampling strategy that reduces LiDAR input size by 95.83\%, enabling efficient policy learning without relying on labeled datasets or manually designed control interfaces. The policy was validated in simulation, achieving a 100\% success rate in straight-row plantations and showing a gradual decline in performance as row curvature increased, tested across varying sinusoidal frequencies and amplitudes.

\end{abstract}

\input{sections/01_Intro}
\input{sections/02_Related_Work}
\input{sections/03_Method}
\input{sections/04_Experiments}
\input{sections/05_Conclusion}
\input{sections/06-Acknowledgment}

\addtolength{\textheight}{-12cm}   

\bibliographystyle{IEEEtran}
\bibliography{ref}

\end{document}

%% file: sections/01_Intro.tex
\section{Introduction}

\begin{figure}
    \centering
    \includegraphics*[width=1.0\linewidth]{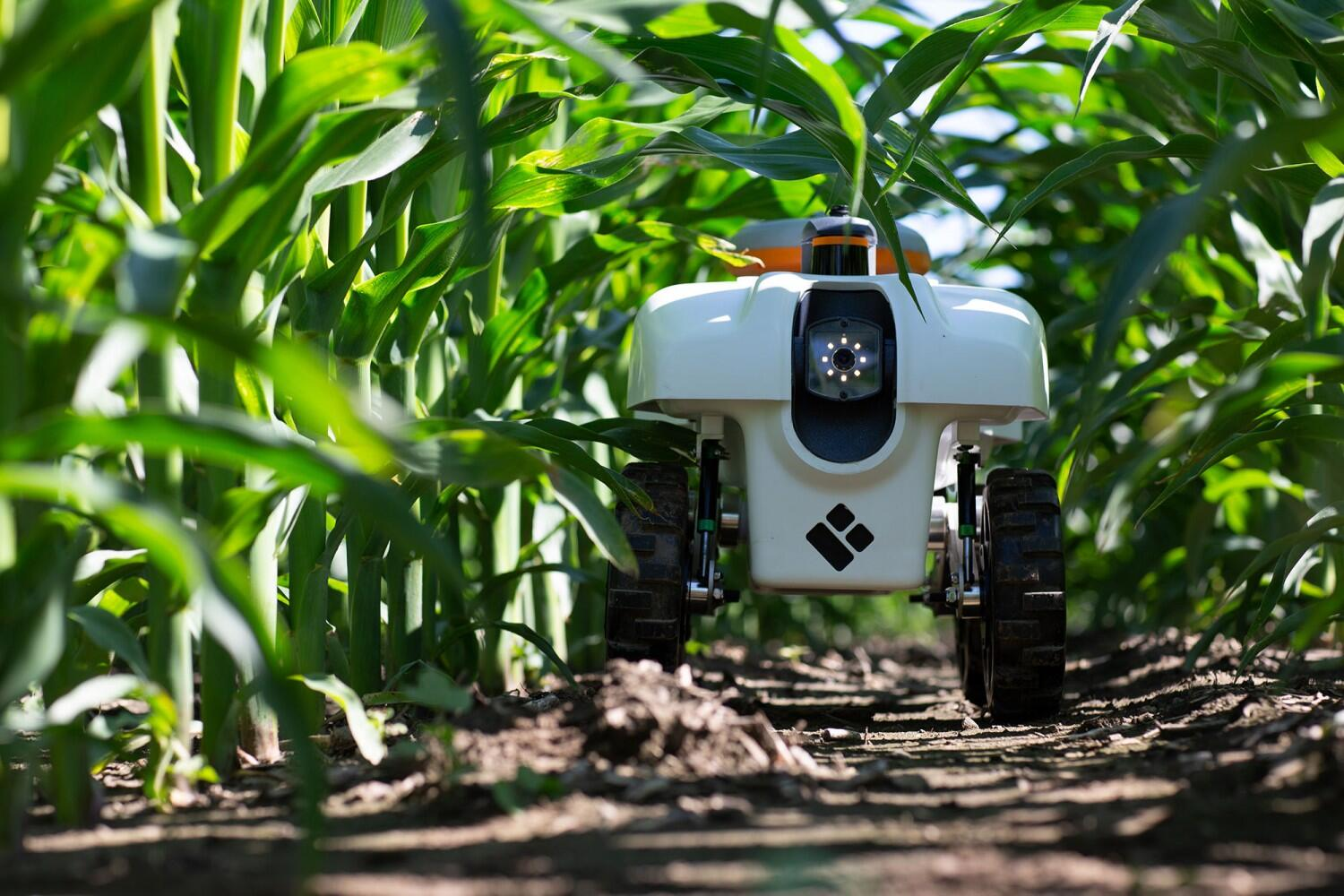}
    \caption{TerraSentia robot navigating in agricultural environments~\cite{terrasentia2020}.}
    \label{fig:terrasentia}
\end{figure}

Autonomous robots have seen significant growth in modern agriculture, particularly for under-canopy tasks such as plant phenotyping, crop row harvesting, and disease scouting. These applications require platforms that are not only compact and agile but also capable of accurately navigating between dense crop rows (Fig.~\ref{fig:terrasentia})~\cite{kayacan2019tracking}. However, reliable navigation in such environments remains an active area of research due to several challenges, including clutter and occlusions caused by narrow row spacing and the high visual variability introduced by different plant growth stages~\cite{debruin2025breaking}.

To enable decision-making systems capable of navigating through plantations in under-canopy environments, robots typically rely on exteroceptive sensors~\cite{gasparino2023cropnav}. This choice is driven by the unreliability of GNSS-based localization in such environments, where signal degradation is common due to foliage occlusion~\cite{sivakumar2024demonstrating}. As a result, there is a need for local perception strategies that allow the robot to infer navigation information from its immediate surroundings.

Recent methods have adopted learning-based approaches that aim to extract row-following features directly from curated and labeled datasets, using cameras or LiDAR sensors as the primary source of perception data. On one hand, cameras provide rich semantic and textural information; however, their performance often degrades in low-light conditions, which are common in dense canopy environments~\cite{bakken2019end, gai2021using}. On the other hand, LiDAR offers accurate depth measurements and is unaffected by lighting variations, making it more robust in shaded or poorly lit areas~\cite{velasquez2022multi, higuti2019under}. Additionally, while a sim-to-real gap exists for LiDAR data, the structural consistency of point clouds tends to generalize well across training scenarios. This characteristic allows learning approaches based on point cloud data to effectively leverage large-scale simulated datasets~\cite{huch2023quantifying}.

In addition, most of these methods typically divide the navigation task into separate modules (e.g., perception, locomotion controller), which introduces challenges in designing effective interfaces between them~\cite{winterhalter2021localization}. As a result, important environmental features may be lost or overly simplified when translated into reference values for the controller. Combined with the high cost and impracticality of collecting labeled data across diverse agricultural conditions, these limitations hinder generalization and reduce overall system robustness. 

For instance, to address the limitations of curated datasets and the challenge of designing interfaces between navigation modules, deep reinforcement learning (RL) can be explored. This approach enables agents to learn end-to-end policies by mapping perceptual observations directly to navigation actions through interaction with the environment. The control policy is optimized via trial-and-error, guided by reward functions that encode the desired task~\cite{baby2024autonomous, huang2021end}.

In this paper, we present an end-to-end row-following navigation system for under-canopy agricultural environments that maps raw 3D LiDAR data directly to control commands using a RL policy trained in simulation. Our approach avoids the need for manually designed interfaces between navigation modules and does not rely on expert-labeled datasets. Additionally, we propose a LiDAR downsampling technique that transforms raw point clouds into compact row maps, significantly reducing the observation space while preserving essential structural features for learning.

The key contributions of this work are:
\begin{itemize}
    \item An end-to-end learning system for crop row-following using only raw 3D LiDAR data.
    \item A voxel-based downsampling technique to convert LiDAR point clouds into compact 2D row maps.
\end{itemize}

%% file: sections/02_Related_Work.tex
\section{Related Work}

\textit{Row Following Navigation} is a critical task in agricultural robotics, enabling autonomous systems to access plantations in both dense under-canopy and open-field environments. Specifically, in under-canopy settings, Higuti et al.~\cite{higuti2019under} introduced the use of LiDAR as a primary perceptive sensor, combined with heuristic-based methods to detect crop rows and estimate the desired steering angle. This angle is then used by a controller to align the robot with the crop row.

Additionally, Velasquez et al.~\cite{velasquez2022multi} proposed a sensor fusion framework that combines LiDAR and IMU data within a Bayesian estimation framework to address the limitations of relying on a single perceptive sensor. This approach enhances local navigation capabilities in environments where GNSS signals are unreliable or unavailable due to interference.

In parallel, despite the work of Sivakumar et al.~\cite{sivakumar2021learned} demonstrating that learning-based approaches using cameras can identify keypoints to trace crop row structures, camera-based perception remains highly sensitive to environmental variability, including changes in lighting and occlusions from dense foliage. In response to these limitations, the work~\cite{pinto2023navigating} extended LiDAR-based perception by applying learning-based models to identify crop rows without relying on hand-crafted heuristics, aiming to achieve more generalizable and robust navigation strategies that do not fail when LiDAR observations deviate from the heuristic bounds.

However, due to the need for large labeled datasets, Affonso et al.~\cite{affonso2025crow} introduced the concept of self-supervised learning for under-canopy navigation. Their approach generates synthetic data by sampling crop row headings, mimicking the behavior of LiDAR readings to train the perception model without manual annotation. While this method reduces the dependency on labeled data, it still faces limitations related to the heuristics used for synthetic data generation and the challenge of interfacing information with the controller.

\textit{Deep Reinforcement Learning} is a decision-making paradigm that enables training end-to-end models to learn optimal actions by maximizing expected rewards through a trial-and-error process, using perceptual observations as input. This approach has demonstrated success across various areas of robotics, particularly in advancing locomotion and navigation systems~\cite{han2024lifelike, cimurs2021goal}. 

Nevertheless, despite its promising potential for under-canopy navigation, RL remains largely unexplored as a navigation controller in this context. This framework enables the direct mapping of raw LiDAR observations to control actions, eliminating the need for manually designed interfaces between perception and control. As a result, it reduces dependence on large, curated, and labeled datasets, allowing the system to autonomously learn task-relevant features through interaction with the environment.

%% file: sections/03_Method.tex
\begin{figure*}
    \centering
    \includegraphics[width=1.0\linewidth]{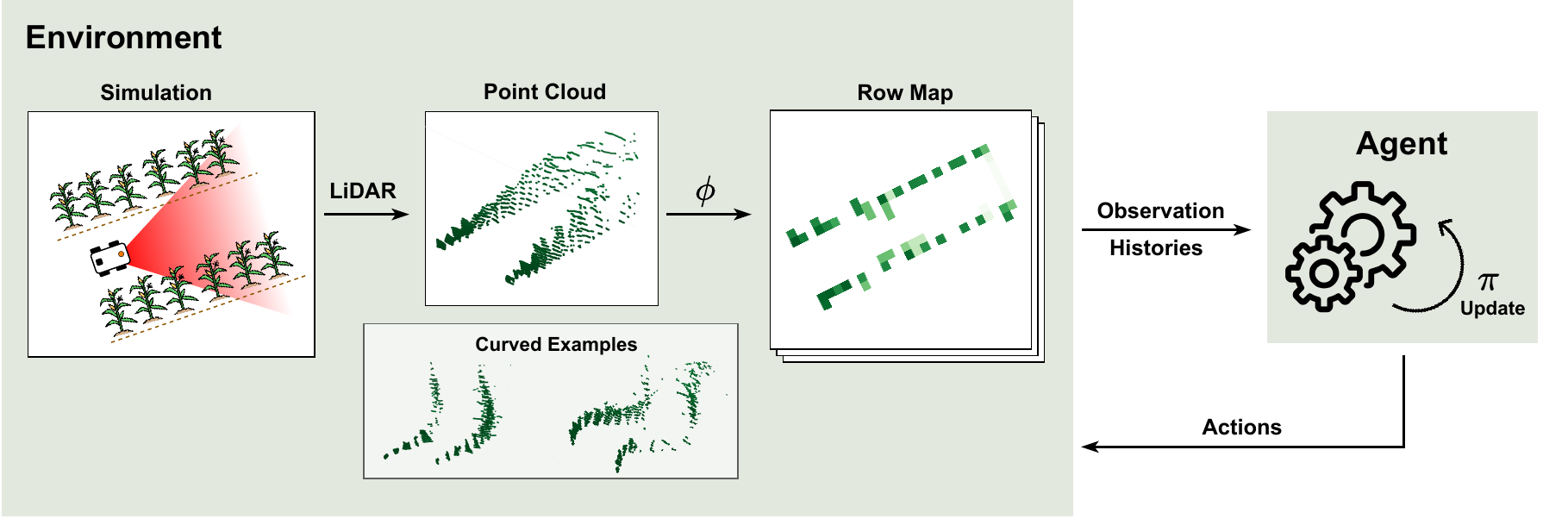}
    \caption{Workflow of the end-to-end row-following navigation system. First, raw LiDAR point clouds captured from the environment are downsampled and transformed into 2D row maps using a transformation function $\phi$, reducing their dimensionality. A finite sequence of row maps is then stacked to construct an observation history, which serves as input to the policy network. Trained using a reinforcement learning algorithm, the policy generates actions that guide the robot to accurately follow crop rows while avoiding collisions.}
    \label{fig:pipeline_overview}
\end{figure*}

\section{Method}

In this section, we present our method for learning an end-to-end row-following navigation policy using only 3D LiDAR data in agricultural environments, based on deep RL. As illustrated in Fig.~\ref{fig:pipeline_overview}, our approach receives raw LiDAR observations from the environment and applies a transformation to reduce the dimensionality of the point cloud, producing a compact representation that serves as the policy’s observation input. The policy is trained to perform the navigation task via a reward function that encourages accurate row following while penalizing undesirable behaviors.

\subsection{Background}

Before presenting our proposed navigation system, we first formulate the problem within the RL framework. Specifically, we model the control task in discrete time at each timestep $k$ as a partially observable Markov decision process (POMDP), since not all system states are directly observable under the constraint of relying solely on 3D LiDAR data.

Moreover, to simplify the problem, prior works have leveraged belief-state approximations constructed from observation histories~\cite{miki2022learning, lee2020learning}. In this work, we approximate the POMDP as a fully observable Markov decision process (MDP) by conditioning the policy on a finite history of LiDAR observations.

The MDP formulation is defined by the tuple $(\mathcal{S}, \mathcal{A}, \mathcal{T}, \mathcal{R})$, where $\mathcal{S}$ is the set of states $s_k$, $\mathcal{A}$ the set of actions $a_k$, $\mathcal{T}(s_{k+1} \mid s_k,  a_k)$ the transition function, and $\mathcal{R}(s_k, a_k)$ the reward function. The objective is to learn a policy $\pi^*$ that maximizes the expected discounted return:

\begin{equation}
    \pi^* = \arg\max_\pi \mathbb{E}\left[ \sum_{k=0}^{\infty} \gamma^k \mathcal{R}(s_k, a_k) \right],
\end{equation}

where $\gamma \in [0, 1]$ is the discount factor.

\subsection{State and Action Space}

Considering the class of wheeled robots equipped with a 3D LiDAR sensor, the state is defined as shown in Eq.~(\ref{eq:state_definion}), based on a history of LiDAR observations. These observations are processed using a transformation that reduces the dimensionality of the point cloud, as described in Subsection~\ref{subsec:data_processing_technique}.
\begin{equation}
s_k = \{ o_{k - \tau + 1}, \dots, o_k \}, \quad o_i = \phi(o_i^{\text{lidar}})
\label{eq:state_definion}
\end{equation}
where, $o_k^{\text{lidar}}$ denotes the raw LiDAR observation, and $\phi(\cdot)$ is a transformation function that reduces its dimensionality to the observation $o_k$. The full state $s_k$ thus comprises a history of $\tau$ processed observations.

Regarding the action space, we adopt a control formulation for a non-holonomic robot. Since the objective of this navigation task is to continuously follow crop rows---without a predefined goal position or terminal state---the forward velocity is kept constant, and the RL policy controls only the angular velocity $\omega$. Consequently, the action variable corresponds to steering the robot to align its heading with the crop row while avoiding collisions.

\subsection{Rewards}

To inform the MDP formulation with the intended task, we designed the reward function to capture the desired behaviors. Inspired by the approach proposed in~\cite{jaeger2025carl}, we define the reward as shown in Eq.~(\ref{eq:reward_function}), following a simple structure that avoids complex multi-objective formulations.

\begin{equation}
\mathcal{R} = r^{\text{task}} \cdot \prod r_i^{\text{penalty}} - r^{\text{collision}}
\label{eq:reward_function}
\end{equation}
where $r^{\text{task}}$ is a dense reward for task completion, $r_i^{\text{penalty}} \in [0, 1]$ are multiplicative penalties discouraging undesirable behaviors, and $r^{\text{collision}}$ is a sparse penalty applied upon collision.

The components used in this work are detailed in Table~\ref{tab:rw_functions}. The task reward is computed from a discretized approximation of the robot’s linear velocity integral, encouraging steady motion along both straight and curved rows.

Penalty terms consist of a sparse penalty for collisions with the plantation and a dense penalty based on the rate of change in the robot’s actions. The latter is clipped to lie within the interval $[0, 1]$, where values closer to 1 result in lower penalization. This promotes smoother control by discouraging abrupt or unnecessary movements and encourages action changes only when necessary to avoid obstacles. The final reward function, which balances task incentives with regularization penalties, is used to train the control policy using the proximal policy optimization (PPO) algorithm~\cite{schulman2017proximal}.

\subsection{LiDAR Data Processing}
\label{subsec:data_processing_technique}

\begin{figure}[b]
    \centering
    \includegraphics*[width=0.82\linewidth]{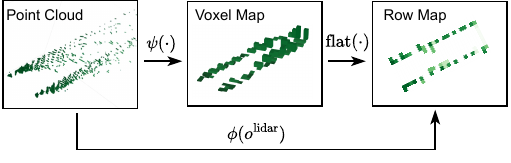}
    \caption{Illustration of the proposed data downsampling method.}
    \label{fig:downsampling_method}
\end{figure}

As the raw 3D LiDAR data has high dimensionality, which poses challenges for efficient processing by neural networks, we propose a method to reduce its complexity, as illustrated in Fig.~\ref{fig:downsampling_method}. First, we define the point cloud with $N$ points as $o^{\text{lidar}} = \{ p_i = (x_i, y_i, z_i) \}_{i=1}^{N}$, and specify a voxel resolution $\delta = (\delta_x, \delta_y, \delta_z)$. 

\begin{equation}
\psi(p_i) = 
(x_i^v, y_i^v, z_i^v)
=\left(
\left\lfloor \frac{x_i}{\delta_x} \right\rfloor,\\
\left\lfloor \frac{y_i}{\delta_y} \right\rfloor,\\
\left\lfloor \frac{z_i}{\delta_z} \right\rfloor
\right)
\label{eq:data_voxelization_process}
\end{equation}

Using this, we discretize the point cloud into a voxel grid via the mapping in Eq.~(\ref{eq:data_voxelization_process}),  resulting in a set of voxelized points $(x_i^v, y_i^v, z_i^v)$, where the superscript $v$ indicates the voxelized representation.

Points that fall into the same voxel are aggregated, thereby reducing redundancy in the representation. However, the resulting voxel grid remains 3D due to the vertical (z-axis) resolution captured by the LiDAR sensor.

To further reduce dimensionality, we flatten the voxel grid along the vertical axis by averaging the occupancy across all $H$ height levels at each $(x, y)$ location, as defined in Eq.~(\ref{eq:data_flattening_process}).
\begin{equation}
    o_{x, y} = \phi(o^{\text{lidar}}_{x, y}) = \frac{1}{H} \sum_{v_i} z_i^v,\, v_i = \psi(p_i) \mid p_i \in {o^{\text{lidar}}_{x,y}}
    \label{eq:data_flattening_process}
\end{equation} 
where the subscript $(x, y)$ denotes the spatial location being computed.

Finally, by applying this downsampling procedure to the original raw point cloud, we obtain a reduced 2D map that preserves the most relevant structural features, making it suitable for use as input to the policy.

\input{tables/tab1_reward_functions}

\setlength\fboxsep{-3.0pt}     
\setlength\fboxrule{0.5pt}     

\begin{figure*}[t]
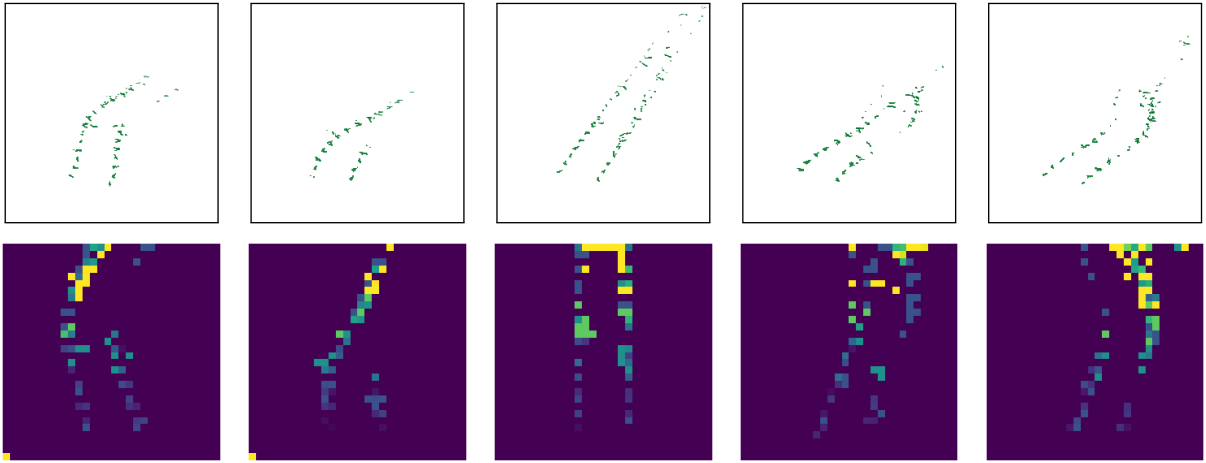

    \centering
    \foreach \i in {0,1,2,3,4}{
        \begin{minipage}[b]{0.17\textwidth}
            \centering
            \fbox{\includegraphics[width=\linewidth]{figures/fig4_pointcloud_\i.png}}
        \end{minipage}
    }

    \vspace{0.2cm}

    \foreach \i in {0,1,2,3,4}{
        \begin{minipage}[b]{0.17\textwidth}
            \centering
            \includegraphics[width=\linewidth]{figures/fig4_occupancy_map_\i.png}
        \end{minipage}
    }
    
    \caption{Representation of the downsampling method. Top: LiDAR point clouds captured over time. Bottom: corresponding row maps used as policy input.}
    \label{fig:downsampling_result}
\end{figure*}

%% file: tables/tab1_reward_functions.tex
\begin{table}[t]
\caption{Reward Functions}
\centering
\begin{tabular}{l l r}
\hline
\textbf{Symbol} & \textbf{Equation} & \textbf{Weight}  \\
\hline
$r^{\text{task}}$ & $ \sum \mathbf{v}$ & 5.0 \\
$r^{\text{penalty}}$ & $\text{clip}\left(1 - \frac{\|\omega_k - \omega_{k-1}\|^2}{\sigma}\right)$ & 1.0 \\
$r^{\text{collision}}$ & $\sum \mathds{1}_{\text{collision}}$ & 1.0 \\
\hline
\end{tabular}
\label{tab:rw_functions}
\end{table}

%% file: sections/04_Experiments.tex
\section{Experimental Results}

In this section, we present the experimental results of our end-to-end crop-following navigation framework driven by a learned policy. We begin by describing the experimental setup, including the TerraSentia robotic platform, the supporting software stack, and the simulation environments used for evaluation. We then analyze the impact of training parameters and benchmark the policy's performance across different plantation patterns. In particular, we evaluate its generalization to previously unseen row geometries and assess its robustness by measuring the maximum distance the robot can travel in simulation without collision.

\subsection{Platform}

For our experiments, we use the TerraSentia robot (EarthSense, Inc.), as shown in Fig.~\ref{fig:terrasentia}. This compact four-wheeled platform is controlled using a learning-based predictive low-level controller~\cite{kayacan2018embedded}. It is equipped with an onboard sensing suite that includes both cameras and a LiDAR, enabling autonomous navigation and data collection in under-canopy agricultural environments. To enhance perception capabilities, the standard platform was upgraded from a 2D LiDAR to a Velodyne 3D LiDAR (VLP-16), providing higher-fidelity data acquisition for decision-making.

\subsection{Simulation Setup}

In this work, we focus our experiments and results on simulated environments. To meet the requirements of supporting RL pipelines and providing high-fidelity physics, we use IsaacSim~\cite{mittal2023orbit}, which offers a GPU-accelerated physics engine along with native integration for RL libraries. As IsaacSim does not natively support the TerraSentia robot or agricultural plantation settings, we developed custom simulation assets to bridge this gap. We accurately modeled the TerraSentia platform, including its joints and actuators, and constructed realistic under-canopy plantation scenarios. These extensions enable end-to-end RL training and evaluation within the IsaacSim framework.

To provide diverse plantation patterns for training and evaluation, we define two row configurations. The first follows a traditional straight crop row layout, while the second introduces a more challenging curved pattern. This curved configuration is modeled using a configurable sinusoidal function, which allows controlled adjustments in amplitude and frequency. Such a setup makes it possible to analyze the robot's ability to maintain row-following behavior under varying curvature levels by assessing its capacity to navigate without collisions along increasingly complex trajectories.

\subsection{Point Cloud Compression}

For the downsampling process, we selected points from the middle four channels of the LiDAR sensor ($H = 4$), where the majority of hits on the plantation are concentrated. These points were voxelized using a cubic resolution of $\delta_{x} = \delta_{y} = \delta_{z} = 0.1$ meters. To focus on local spatial context relevant for capturing the plantation's row geometry, we extracted a fixed region of interest spanning 30 voxels along both the $x$- and $y$-axes. This reduced the original input of 7,200 3D points to a compact 30 × 30 row map (900 cells), achieving a 95.83\% reduction in data.  The resulting representation balances dimensionality reduction with the preservation of meaningful structural information for learning.

Figure~\ref{fig:downsampling_result} illustrates examples of the raw LiDAR point cloud alongside the corresponding row map, providing a visual reference for the policy input. These examples highlight how the map representation captures the plantation structure across different stages during navigation, including both straight and curved scenarios. Notably, due to the angular behavior of the LiDAR beams, it is also common to observe a higher density of points on sections of the plantation that are farther from the robot.

\begin{figure*}[t]
    \centering
     \includegraphics[width=1.0\linewidth]{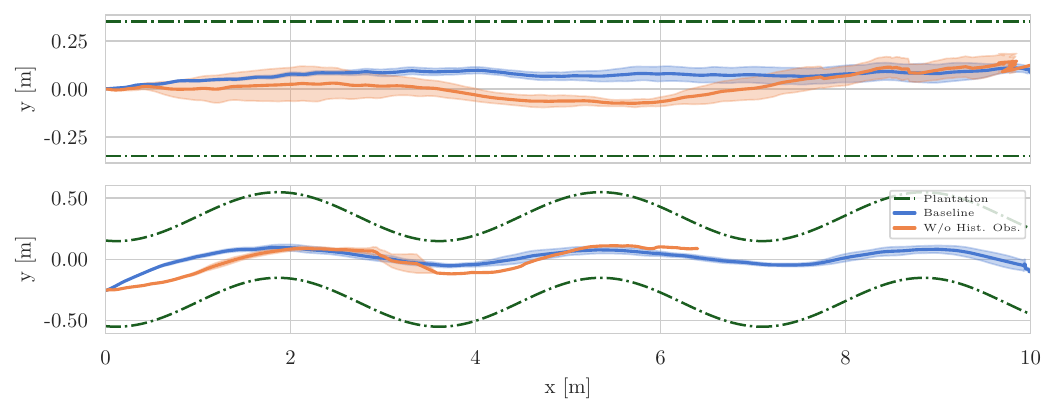}
    \caption{Navigation trajectories of the trained policies bounded by plantation geometry. Top: Straight-row scenarios. Bottom: Curved-row scenarios.}
    \label{fig:navigation_trajectories}
\end{figure*}

\subsection{Policy Training}

Before presenting the validation of the trained policy in downstream tasks, we first report the numerical results related to training performance in terms of episode returns. 

Training was performed using the \texttt{rsl-rl} library~\cite{schwarke2025rsl} with PPO hyperparameters listed in Table~\ref{tab:ppo_parameters}. Policies were trained in a sinusoidal plantation environment (frequency 1.8 Hz, amplitude 0.20 m), providing a moderately challenging row-following scenario.  A history length of $\tau = 3$ was used, and training was performed under three different settings:
\begin{itemize}
\item Policy with all features (baseline);
\item Policy without history observations;
\item Policy without downsampling.
\end{itemize}

\begin{figure}[h]
    \centering
    \includegraphics[width=1.0\linewidth]{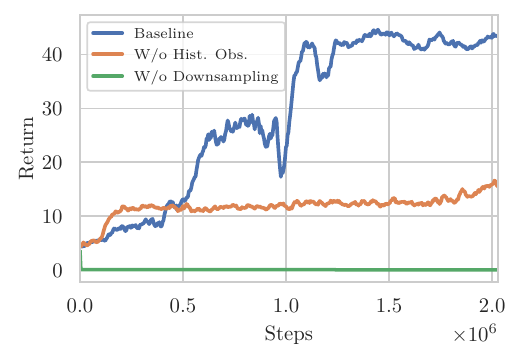}
    \caption{Average episode return of each policy.}    
    \label{fig:training_policies}
\end{figure}

Figure~\ref{fig:training_policies} illustrates the training returns over environment steps for the different policy configurations. These comparisons highlight the influence of temporal context and input size on learning performance and policy robustness. Without downsampling, the high-dimensional input overwhelms the network, leading to failed learning. A single row map allows learning to begin, but including a history of observations significantly improves performance. This supports the POMDP formulation, where temporal context helps address partial observability.

\input{tables/tab2_ppo_parameters}

\subsection{Navigation Trajectory}

After presenting the results of policy training and the downsampling visualization, we now analyze the trajectories produced by the stochastic navigation policies in both straight and curved plantation scenarios---corresponding to the training environment settings.

Figure~\ref{fig:navigation_trajectories} presents the trajectories executed over ten trials along a 10-meter crop row for the trained policies. The policy trained without downsampling is excluded, as it failed to learn a viable behavior.  The results show that the proposed reward formulation promotes navigation that stays near the centerline of the row, despite the absence of an explicit reward for lateral deviation.

Comparing the two policies, both can complete the straight-row scenario; however, the version without historical observations tends to drift further from the centerline. In the curved-row scenario, only the baseline policy successfully completes the track, while the no-history policy averages just 6.3 meters before failure. These results also support our POMDP formulation, demonstrating that access to a sequence of observations provides crucial temporal context that improves the policy’s ability to infer and adapt to the navigation task profile.

\input{tables/tab3_results_experiments}

\subsection{Generalization to Unseen Scenarios}

As a final experiment, we evaluate the robustness of the proposed method by analyzing its performance across a variety of plantation patterns. These include a straight configuration and a series of sinusoidal curves generated by systematically increasing the frequency and amplitude of the plantation rows. Table~\ref{tab:results_experiments} summarizes the results for all configurations, considering the task of navigating through a 100-meter row without collisions over 15 trials.

\begin{figure}[b]
    \centering
    \includegraphics[width=1.0\linewidth]{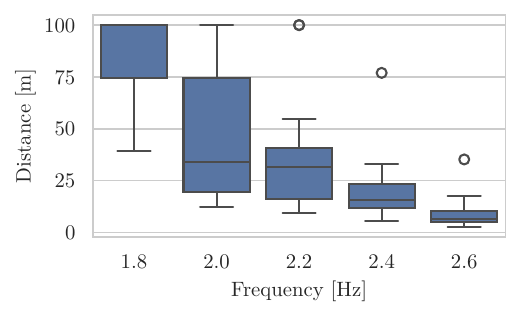}
    \caption{Distance navigated at different sinusoidal row frequencies with fixed amplitude.}    
    \label{fig:box_plot_experiments_frequency}
\end{figure}

In the baseline configurations---i.e., the straight plantation and the curved pattern used during policy training---the navigation performance was satisfactory.  In the straight-row scenario, the agent consistently completed the full 100-meter row, achieving a 100\% success rate. In the curved training scenario, it achieved a mean and standard deviation of $83.89 \pm 22.72$ meters. It is worth noting that real-world plantation layouts typically resemble the straight-row pattern, highlighting the practicality of the trained policy for real-world deployment.

To evaluate the generalization capability of the policy, we conducted additional tests with increased terrain complexity:
\begin{itemize}
\item Increasing the frequency, keeping the amplitude fixed;
\item Increasing the amplitude, keeping the frequency fixed.
\end{itemize}

\begin{figure}[b]
    \centering
    \includegraphics[width=1.0\linewidth]{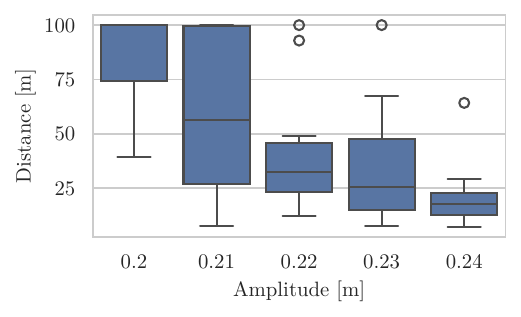}
    \caption{Distance navigated at different sinusoidal row amplitudes with fixed frequency.}
    \label{fig:box_plot_experiments_ampliture}
\end{figure}

In both cases, we observed a gradual decline in performance as either the frequency or the amplitude of the curves increased. Figures~\ref{fig:box_plot_experiments_frequency} and~\ref{fig:box_plot_experiments_ampliture} show that the policy achieved shorter navigation distances and lower success rates for full row completion as the terrain became more complex. This trend is consistent with the increased number of collisions observed in earlier stages.

In simpler configurations, the results exhibited higher variance because in some trials the policy was able to complete the task, despite struggling to do so consistently. This behavior suggests that the policy can operate on patterns that differ from those encountered during training, although its performance is less reliable under such conditions. Finally, the average time per trial followed a similar pattern to the navigated distance. Trials that ended prematurely due to collisions resulted in shorter execution times.

This reduction in performance can be explained by the difficulty RL policies face when operating in conditions that differ significantly from the training distribution. As the curves became more extreme, the controller encountered scenarios that required more abrupt changes in direction and lower linear velocities, which were not represented during training.

%% file: tables/tab2_ppo_parameters.tex
\begin{table}[h]
\caption{PPO Training Hyperparameters}
\centering
\begin{tabular}{l l}
\hline
\textbf{Parameter} & \textbf{Value} \\
\hline
Clip ratio ($\epsilon$) & 0.2 \\
Learning rate & $10^{-3}$ \\
Discount factor ($\gamma$) & 0.99 \\
GAE-lambda & 0.95 \\
Desired KL divergence & 0.01 \\
Entropy coefficient & 0.005 \\
Value loss coefficient & 1.0 \\
Batch size &  4096\\
Actor hidden dims & [512, 256, 128] \\
Critic hidden dims & [512, 256, 128] \\
\hline
\end{tabular}
\label{tab:ppo_parameters}
\end{table}

%% file: tables/tab3_results_experiments.tex
\begin{table*}[h]
\caption{Evaluation of the trained policy over different plantations}
\centering
\begin{tabular}{cccccc}
\toprule
\textbf{Frequency (Hz)} & \textbf{Amplitude (m)} & \textbf{Avg Distance (m)} & \textbf{Std Distance (m)} & \textbf{Avg Time (s)} & \textbf{Std Time (s)} \\
\midrule
\multicolumn{6}{c}{(i) Baseline configurations} \\
0 & 0 & 100 & 0 & 177.49 & 0.56\\
1.8 & 0.20 & 83.89 & 22.72 & 149.16 & 40.33 \\
\midrule
\multicolumn{6}{c}{(ii) Fixed amplitude of 0.20 m and varying frequency} \\
2.0 & 0.20 & 44.99 & 32.09 & 80.31 & 56.99 \\
2.2 & 0.20 & 36.87 & 28.57 & 66.19 & 50.79 \\
2.4 & 0.20 & 21.05 & 17.50 & 38.26 & 31.55 \\
2.6 & 0.20 & 9.46  & 8.30  & 17.41 & 15.21 \\
\midrule
\multicolumn{6}{c}{(iii) Fixed frequency of 1.8 Hz and varying amplitude} \\
1.8 & 0.21 & 56.32 & 36.09 & 100.63 & 64.20 \\
1.8 & 0.22 & 38.78 & 26.19 & 69.29  & 46.36 \\
1.8 & 0.23 & 34.65 & 26.58 & 62.26  & 47.53 \\
1.8 & 0.24 & 19.57 & 14.00 & 35.67  & 25.57 \\
\bottomrule
\end{tabular}
\label{tab:results_experiments}
\end{table*}

%% file: sections/05_Conclusion.tex
\section{Conclusion}

We presented an end-to-end row-following navigation system for robots operating in under-canopy agricultural environments, based on RL within a POMDP framework. The system relies solely on 3D LiDAR input, which is transformed into a compact row map through a voxel-based downsampling method. A sequence of these compact observations is used as input to a policy network trained with a reward function that encourages forward progress while penalizing collisions and abrupt changes in control actions.

The proposed approach was evaluated entirely in simulation and demonstrated reliable performance across a range of scenarios. In straight-row plantations, the system consistently completed 100-meter navigation tasks. Furthermore, we assessed the policy under increasingly complex curved-row configurations by varying the frequency and amplitude of sinusoidal paths. The system exhibited a gradual decline in performance, indicating robust generalization within the limits of the training distribution.

For future work, we plan to extend this work by incorporating a latent perception representation to enhance policy generalization across diverse plantation layouts. We also intend to expand the range of training layouts through domain randomization of factors such as plant size, spacing, density, and shape. Additionally, we aim to validate our system and future improvements in real-world environments.

%% file: sections/06-Acknowledgment.tex
\section*{Acknowledgment}

This work was supported by São Paulo Research Foundation (FAPESP) grant no. 2024/09535-9, Brazilian National Research Council (CNPq) grants no. 308092/2020-1, and Petrobras, using resources from the R\&D clause of the ANP, in partnership with the University of São Paulo and the intervening foundation FAFQ, under Cooperation Agreement No. 2023/00016-6 e 2023/00013-7.